# Connecting Visual Experiences using Max-flow Network with Application to Visual Localization

A.H. Abdul Hafez[1*], Nakul Agarwal[2*], and C.V. Jawahar[3]

[1]Hasan Kalyoncu University    [2]University of California, Merced    [3]CVIT, IIIT Hyderabad

*Abstract*— We are motivated by the fact that multiple representations of the environment are required to stand for the changes in appearance with time and for changes that appear in a cyclic manner. These changes are, for example, from day to night time, and from day to day across seasons. In such situations, the robot visits the same routes multiple times and collects different appearances of it. Multiple visual experiences usually find robotic vision applications like visual localization, mapping, place recognition, and autonomous navigation. The novelty in this paper is an algorithm that connects multiple visual experiences via aligning multiple image sequences. This problem is solved by finding the maximum flow in a directed graph flow-network, whose vertices represent the matches between frames in the test and reference sequences. Edges of the graph represent the cost of these matches. The problem of finding the best match is reduced to finding the minimum-cut surface, which is solved as a maximum flow network problem. Application to visual localization is considered in this paper to show the effectiveness of the proposed multiple image sequence alignment method, without loosing its generality.

Experimental evaluations show that the precision of sequence matching is improved by considering multiple visual sequences for the same route, and that the method performs favorably against state-of-the-art single representation methods like SeqSLAM and ABLE-M.

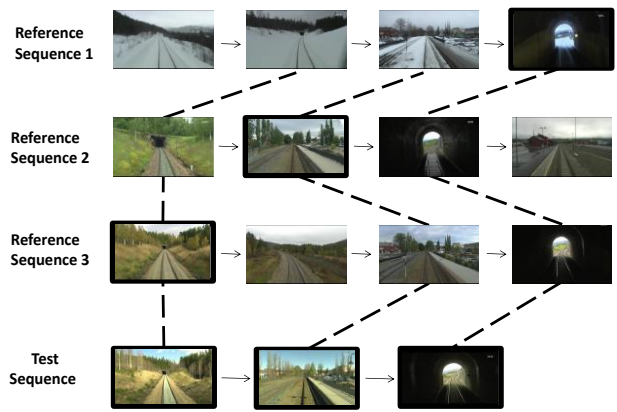

Fig. 1. Connecting new visual experience (test image sequence) to multiple previous visual experiences (reference sequences.) Dashed line segments connect a frame from the new visual experience to its corresponding best match from each previous visual experience. Frames that correspond to best global match are represented with bold outlines.

## I. INTRODUCTION

Visual localization and mapping in outdoor urban environments is challenging and still an open problem [1], [2]. The appearance of our environment may change due to several reasons like perceptual aliasing, dynamic elements in the environment, and object and camera occlusion. Indeed, our observations may change drastically in appearance from day to night or summer to winter for example. In addition to changing appearance in time, places may change in a cyclic manner [3]. Therefore, it is important to hold multiple representations rather than a single description. These representations are stored in the form of visual experiences, *i.e.* visual odometry and features information [4], [5], [6], [7] or more commonly a sequence of images [8], [9], [10], [11].

Several works in the literature have discussed the need for multiple representations and life-long visual localization [12], [4], [13], [14], [15]. Churchill and Newman proposed in [4] to store a new visual experience whenever the localizer is not able to localize the current stream with one of the previously available experiences. Ranganathan *et al.* has pointed in [16] that for good visual localization

*  denotes equal contribution

over night and day time, around three or four images are required for every place. However, storing multiple visual experiences of a certain place/route brings another challenge. This challenge is the difficulty in localizing the robot across these different experiences. This problem can be solved by connecting these multiple experiences. A visual experience is usually represented by a graph that contains a set of nodes each of which represents a location and a set of directed edges that connect them in order. It was suggested in [4] to connect nodes from different experiences and represent the same location by an edge from one experience to another.

Connecting experiences allows us to fully exploit the given sequences, this reduces the need for more new experiences. In addition, it allows sharing the same information like geometric transformations or the GPS annotation. The connectivity of the stored experiences was proposed in [4], [9], where two main types of connection discoveries were presented. One is the live discovery which happens when a localizer matches the curent stream to more than one visual experiences at the same time. This is used as an indication that the current node in both experiences represents the same place. The second is a GPS-based discovery in which GPS tags are recorded in synchronization with recording the visual experiences. These GPS readings of newly recorded experience are used to connect its nodes to previous experiences. This method however, is subject to errors in GPS readings and in the frame/node annotation. Appearance based visual localization methods [8], [17], [18] connect nodes from different experiences to each other based on similarity in visual appearance. The work presented in [19] exploits combination of appearance similarity and GPS priors about the environment. All these appearance based methods formulate the connectivity discovery as a sequence to sequence matching without any consideration for the multiple representation requirements of the environment.

In this paper, we address the problem of connecting visual experiences using a multiple image sequences matching framework. We assume that several representations of the environment are available in the form of visual experiences, i.e. a set of image sequences or a sequence of extracted features without loosing generality. The proposed method aligns a test sequence of frames to multiple reference sequences. The problem can be handled as alignment of multiple reference sequences as well, but the former formulation is selected to be compatible with the context of visual localization as in [4], [18], [8], [10], i.e. our selected application. For example, the visual experiences are recorded sequentially in [4], [18]. Every time a new visual experience is recorded for the same place, it is applied as input to the alignment algorithm to connect it with the previously available experiences. In the initial situation when only one previous experience is available, the matching is done as one sequence to one sequence.

The contribution of this paper is summarized in Fig. 1. It depicts an algorithm that connects new visual experience to previous visual experiences. As shown in Fig. 1, the proposed algorithm maps each frame in the test sequence to one frame from each of the reference sequences. A matching graph network is built in which every vertex (or node) represents a possible match between a given test frame and a frame from one of the reference sequences. The graph is considered a *maximum flow network* with source and sink vertices added to the graph. The best possible match is determined by optimizing the flow via the network between the source and sink vertices. The maximum flow is represented by a *min-cut surface* that separates the source and sink. This surface, in fact, is the surface of best matches between each frame from the test sequence and each of the reference sequences.

The remaining of the paper is organized as follows. Section II reviews the most related and recent works on visual localization using multiple experiences, image sequence alignment for visual localisation, and works that utilise network flow for image matching. Section III introduces the problem of alignment of two image sequences, and then it shows the generalization of the problem to alignment of multiple sequences. After that, multiple image sequence alignment is presented as maximum-flow network problem in Section IV. Section V presents experimental evaluations of the proposed multiple image sequence alignment method with application to visual localization problem.

## II. RELATED WORK

Using multiple visual representations and experiences of the same environment are now common in the literature. These are similar to using video sequences for ground vehicle. Learning and probabilistic methods for visual localization [20], [21], [22], [23], [24], [5], [10] presents a framework to learn the feature observability and a priori models, then uses them to predict the location of the system. For example, FabMap localization algorithm [20], [21] treats the multiple sequences as a loop closure problem. Life-long robotic mapping by learning the temporal co-observability between different places is presented in [22]. The learning processes use video sequences recorded over long period of time for selected place locations. A summary map is proposed in [5] as a solution for mapping from multiple image sequences problem, in which scoring functions are used, similarly to [10], [25], to update the map after every arrival of new experience. Multiple visual representations are also considered in [23]. Here, a probabilistic cost map is created in a self-supervised manner using a Gaussian process.

Visual localization in crowded environment from multiple experiences was proposed in [10], [25] and generative methods for long-term place recognition was proposed in [24]. These works address the case where the vehicle frequently visits the same environment, updating its *a priori* knowledge after each visit. The aim was to learn useful and visually stable features that are static in the environment. As further training image sequences are obtained for each place, the feature descriptions are updated over time in such a way that the model filters the dynamic effects in the scene. Long-term localization is achieved based on visual experience as presented in [4], [9]. The visual experiences here are stored as a series of visual odometry data attached to a sequence of visual images. The experiences are connected to each other via GPS priors or geometric consistency among image frames. However, there is no systematic method proposed in these papers for establishing such connections between inter-experiences frames.

Several recent works formulate the problem as image sequence matching [8], [26], [17] and [19], [27], [18]. In these works, a test image sequence is matched against previously visited (learnt) reference sequence. SeqSLAM [8] is a pioneering work in this category. It searches for all possible matches in the complete reference sequence in a sequence to sequence alignment framework. A cost matrix $C$ is established in which a path that has minimum cost is found to represent the best matches.

In the above methods, there is no explicit frame to frame alignment for the multiple image sequences available to the system. This alignment may help for finding direct connections between places that show different appearances in different sequences, like the ones required between inter-experiences in [4], [9]. In contrast, we approach the problem of connecting visual experiences for localization purpose as a multiple image sequences alignment, where the test or current sequence is aligned with all previously available reference sequences.

In the context of visual localization via sequence-to-sequence matching using flow network, we find those which are published in [28], [19], [18] most relevant. They address localization by achieving image sequence matching problem solved as minimum cost path in 2D flow network. Flownets are directed graphs in which a flow traverses from a source node to a sink node. Each of the other nodes represent a match between test and reference frames in the cost matrix. The maximum flow represents a minimal cost path in the cost matrix, which is the best frame-to-frame test and reference sequences matching. To allow creation of path even in case there are no matches, hidden nodes are proposed. This is the situation where different velocity profiles need to be considered. In contrast, we propose to use the flow network to achieve multiple image sequence matching. Our formulation is to find a *maximum flow network* or the *min-cut surface* in a 3D flow network. The proposed formulation and how it is different from the 2D cited above are presented in the next two sections. It is worth to note here that the framework proposed in [18] aims at achieving visual localization while our proposal aims at connecting visual experiences via alignment of image sequences, which in turn facilitates the localization process later. This is why we've selected our experiments in the context of visual localization.

### III. IMAGE SEQUENCE ALIGNMENT

#### A. Features and matching criterion

Matching images of the same place across seasons and under pose changes is a challenging task. The appearance of the same place is likely to change substantially with course of time, be it across seasons, months, days or sometimes even hours. Such changes have a large impact on the performance of the matching process. For example, during a snowy winter day, the landscape is covered by snow and the scene is poorly illuminated, yielding few texture and low contrast. Image matching methods that make use of keypoint-based feature detectors to compute region descriptors [29] are likely to detect only a small number of keypoints in such cases. The proposed alignment framework can be used with any of the recent discriminating features like SIFT [30], [31], BRIEF [32], sequence of binary features [27], HOG [33], or deep CNN features [12]. Histogram of Oriented Gradients (HOG) features are proved to be hardly discriminating over temporal environment changes. It is widely common in the recent literature now that HOG features outperform traditional features like SIFT, SURF and Harris in different aspects, particularly time complexity and discriminability [34], [35], [36], [37].

We make use of HOG descriptors in our experiments and then compare with CNN features extracted from deep residual networks [38] as well. This comparison shows the applicability of the proposed method to other kinds of features. We use HOG as proposed in [33]. We use grid cells of $30 \times 30$ pixels for a $960 \times 540$ image $I$. The 50 bin histograms of gradients calculated from each of the cells are concatenated in $1D$ feature vector to form the image feature descriptor $\mathbf{A}_I$ of the image $I$. The matching cost between a frame $x_j$ from the test sequence and frame $y_i$ from the reference sequence is calculated using the cosine function between the vectors that represent their respective HOG feature descriptors $\mathbf{A}_{x_j}$ and $\mathbf{A}_{y_i}$. This cost is given by

$$\text{cost}(u) = 1 - \cos(\mathbf{A}_{x_j}, \mathbf{A}_{y_i}) = 1 - \frac{\mathbf{A}_{x_j}^T \cdot \mathbf{A}_{y_i}}{|\mathbf{A}_{x_j}||\mathbf{A}_{y_i}|}, \quad (1)$$

where $u$ is the possible match between the two frames $x_j$ and $y_i$. A better match would have a lower cost. The cost selected in this way is proportional to the angle between the the two feature vectors, which is more important than the magnitude. The magnitude in HOG features represents the frequency of a specific orientation of the gradient. Since changes in lightning and appearance may affect the magnitude more than the orientation, we care about the direction of the feature vectors instead of the magnitude. This is achieved by using the cosine cost function instead of, for example, the Euclidean one.

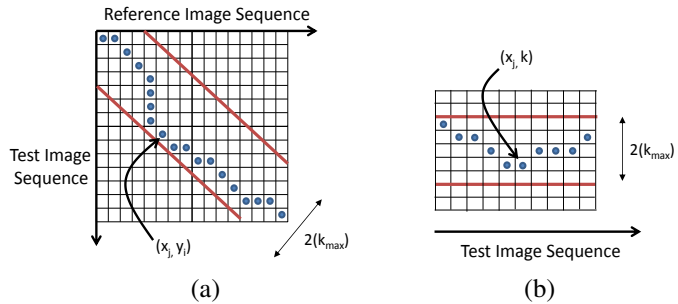

Fig. 2. A test image sequence is matched to a reference sequence for visual localization. In (a), matching matrix $C$ and the path of best matches $(x_j, y_i)$ as blue circles are shown. In (b), the best matches are represented using the couple $(x_j, k)$, without direct appearance of reference sequence. The value of $k$ lies within the range $2k_{max}$.

#### B. Two sequence matching

The possible best matches between the two sequences form a path in the matching matrix $C$ [8], [28], [19], as shown in Fig. 2(a). Every frame $x_j$ from the test sequence matches a frame $y_i$ from the reference frame. It is easy to represent the match $(x_j, y_i)$ in the matching path as $(x_j, x_j + k)$, where $k = y_i - x_j$ is the index *shift* between the test frame and the matched reference frame. This shift is due to differences in velocity profiles. The value of $k$ lies in the range $[-k_{max}, +k_{max}]$, where $k_{max}$ is the maximum possible shift between two matched frames, one from the test and another from reference sequence as shown in Fig. 2(a). We assume in this work that the value of $k_{max}$ is predictable, while leaving the investigation on a method that estimate it for future work. It can be selected in the worst to half the length of the shortest reference sequence.

The matching matrix between test sequence $x_j$ and reference sequence $y_i$ can be modified into the equivalent compact formulation in Fig. 2(b), where only the test sequence appears directly. In that case, each potential match for a particular test image lies in the range $[-k_{max}, +k_{max}]$ around it, and is represented in the form $(x_j, k)$. Here, $x_j$ represents a frame from the test sequence, and $k$ represents the index shift.

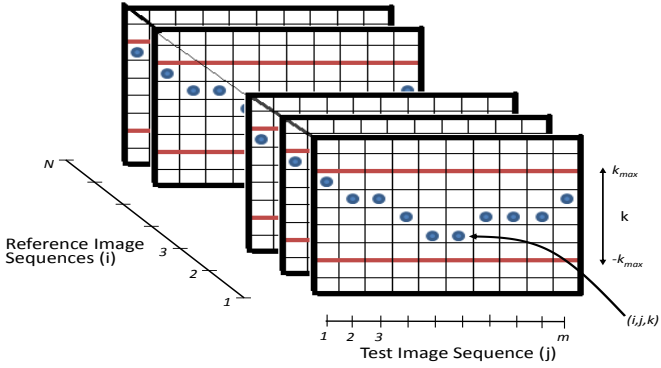

Fig. 3. Matching across multiple reference sequences. The 2D representations $(x_j, k)$ of matching a test sequence with several reference sequences are concatenated in layers so that test sequence is matched with all reference sequences.

### C. Complete matching over multiple sequences

The 2D network representation $(x_j, k)$ aligns one test sequence with one reference sequence of frames. This representation can be extended to align one test sequence with multiple reference sequences for visual localization.

This extension is depicted in Fig. 3, where all the matching reference sequences are lined up together, resulting in the formation of a 3D network. All matching paths, each of which defines the matching of test sequence with one of the reference sequences, are now assembled to form a *matching surface*. This surface contains local matches for every frame from the test sequence with a frame from each of the reference sequences. One thing to note here is that this 3D network is 4-connect since matching with each reference sequence is done independently. In the next section, we present an algorithm that finds the matching surface, along with a global best match, by transforming the network into a 6-connect 3D network, which takes in consideration the spatial relationship between reference sequences as well. Traditional sequence matching methods like Dynamic Programming and Hidden Markov Models used in [39], [17] are not suitable to this matching scheme. We propose to solve the problem as a Maximum-Flow problem [40] in 6-connect 3D network with $k$ as the third dimension as explained in the next section.

## IV. MAX-FLOW NETWORK FOR MULTIPLE IMAGE SEQUENCE ALIGNMENT

We show in this section how connecting visual experiences can be achieved with respect to an environment represented by multiple reference image sequences. The problem is formulated as alignment of a sequence of test frames $\{x_j\}$ to a set of reference sequences $Y_i$, where $i = 1, \cdots, N$, where $N$ is the number of reference sequences. A frame $y_{i,j}$ is the $j$th frame in the sequence $Y_i$. The number of frames in the sequence $Y_i$ is $m_i$ and the number of frames in $\{x_j\}$ is $m$. We assume that all sequences are recorded by a camera mounted on a moving vehicle or robot, and locations are visited sequentially. The vehicle traverses the same route multiple times. Every run is represented by a reference sequence of images $Y_i$. Of course, the robot or vehicle moves with different speeds and may stop several times at different places like traffic lights. However, we assume that matches between different sequences can be found within the range $[-k_{max}, +k_{max}]$.

We propose to use 3D flow network (directed graph $G = (V, E)$ ) to find the *matching surface* by solving a maximum flow network problem, see Fig. 4(a). Matching surface contains a best matching frame from each of the reference sequences that corresponds to every test frame $x_j$ from the test sequence. The architecture of the directed graph is explained in the following subsections. It is also shown that the matching surface is nothing but the min-cut surface of the graph $G = (V, E)$, which results from cutting the saturated edges when a maximum quantity of flow $F$ is running through the network. It can be also considered as the problem of finding the min-cut of the graph $G$, which is represented by the set of saturated edges that have minimal capacities.

### A. Nodes

Let us define a 3D mesh formed by directed graph $G = (V, E)$. The set of vertices $V$ is defined as

$$V = \{s, (i, j, k), t\},$$

where $s$ and $t$ are called the source and sink vertices respectively. Here $j \in [1, \cdots, m], i \in [1, \cdots, N], k \in [-k_{max}, \cdots, k_{max}]$, and $m$ is the maximum sequence's length.

The architecture of the directed graph is represented in Fig. 4(a). The vertex $(i, j, k)$ represents a match between the $j$th frame from the test sequence and the $(j + k)$th frame from the $i$th training sequence. Note that $i, j \in \mathcal{N}$ and $k \in \mathcal{I}$, where $\mathcal{N}$ and $\mathcal{I}$ represent natural numbers and integers respectively. Indeed, the distance between any two neighbor vertices is $\| u - v \| = 1$ when they are next to each other from the same sequence or two frames from different consecutive sequences with the same order. The problem now is to find the maximum quantity of flow $F$ from source vertex $s$ to sink vertex $t$. This maximum depends mainly on the capacities of the edges $E$ which are presented in the next subsection.

### B. Edges

All nodes except the nodes on the boundaries in the graph $G = (V, E)$ are 6-connected, with three edges originating from the node and three ending on it. The 6-connected architecture of the different edges around one example node is shown in Fig. 4(b). The set of edges $(u, v)$ that interconnect the mesh $\{(i, j, k)\}$ includes the following edges

$$(u, v) = \begin{cases} ((i, j, k), (i, j, k+1)), \\ ((i, j, k), (i, j+1, k)), \\ ((i, j, k), (i+1, j, k)). \end{cases} \quad (2)$$

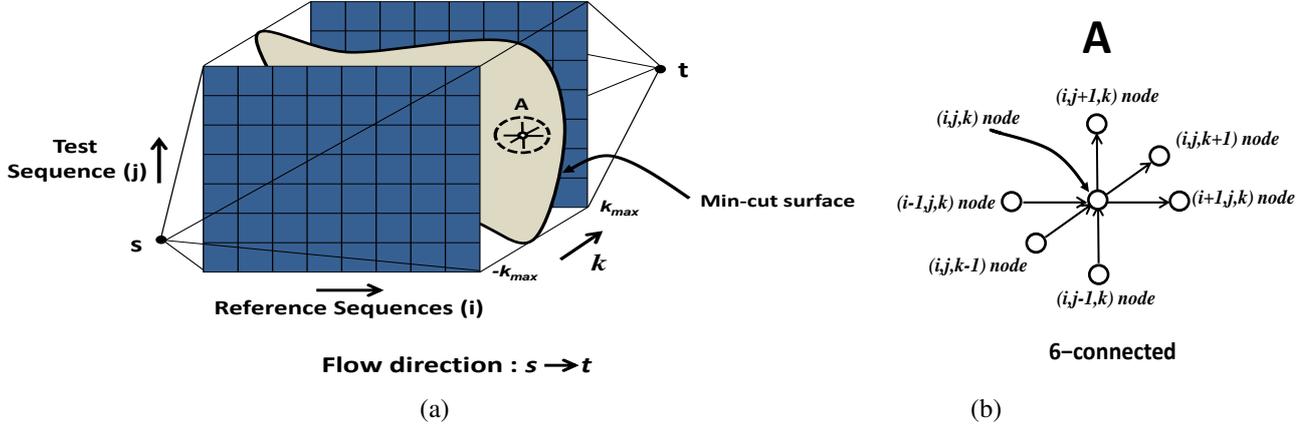

Fig. 4. Matching across multiple reference sequences as a maximum flow problem. The architecture of the 3D flow network and the min-cut surface is shown in (a). Here, $j$ is the index of the test frames and and $i$ is the index of the reference sequences. In (b), the 6-connected edge structure for a selected node example **A** is depicted. The flow direction is from $s$ to $t$ incrementally along the direction of $k$.

These edges describe the spatial relationship between adjacent frames along adjacent sequences as well. They connect adjacent nodes as can be seen in Fig 4(b).

The edge $((i,j,k),(i,j,k+1))$ is aligned with the direction of flow. It represents the closeness of the match of the $j$th frame from test sequence with the $(j+k)$th frame and $(j+(k+1))$th frame from $i$th training sequence. It checks for sequential nature of both the same reference sequence and the test frame. We call it here onward the *shift edge*. Its capacity is denoted by $c_{sh}(u,v)$.

The edge $((i,j,k),(i,j+1,k))$ is perpendicular to the direction of flow in general. It represents the closeness of the match of the $j$th frame from test sequence with the $(j+k)$th frame from $i$th reference sequence. It also presents the closeness of the $(j+1)$th frame from test sequence and $((j+1)+k)$th frame from $i$th training sequence. It checks for sequential nature within the same training sequence with different test frames. The edge $((i,j,k),(i+1,j,k))$ is also perpendicular to the direction of flow in general. It represents the closeness of the match of the $j$th frame from test sequence and the $(j+k)$th frame from $i$th training sequence as well as the $(i+1)$th training sequence. It checks for sequential nature for same test frame within different training sequences. This edge is a reflection of our basic assumption that there is some degree of synchronization among the different training sequences. The last two kinds of edges are called *occlusion edges*. Their capacity is denoted by $c_{oc}(u,v)$.

In addition to the three types of edges mentioned above, we have the following two kinds of edges:

$$(u,v) = \begin{cases} (s,(i,j,-k_{max})), \\ ((i,j,k_{max}),t) \end{cases} \quad (3)$$

These edges connect the mesh $\{(i,j,k)\}$ to the source $s$ and the sink $t$ respectively.

*C. Edge's capacity*

Every edge of $E$ is associated with a capacity value that limits the flow through it. The capacity of edge $c(u,v)$ between any vertices $u$ and $v$ is directly proportional to the cost of matching them. The less matching cost results with less edge capacity. Edges with lower capacities are more likely to be saturated by the maximum flow. Inversely, edges with higher capacities are unlikely to be saturated.

Since every vertex in the graph is a possible match, the edge capacity is computed from the matching costs of the two vertices connected by the edge. The capacity of *shift edges* can be calculated, for example, as follows:

$$c_{sh}(u,v) = \frac{cost(u) + cost(v)}{2}, \quad (4)$$

where $cost(u)$ is the cost of the match that corresponds to the node $u$ in Eq. (1). For example, if $u = (i,j,k)$, then $cost(u)$ is the cost of matching the $j$th frame from the test sequence to the frame $j+k$ from the training sequence $i$.

The capacity of *occlusion edge* is given as

$$c_{oc}(u,v) = \eta \cdot c_{sh}(u,v) \text{ where } (0 \leq \eta \leq \infty), \quad (5)$$

where $\eta$ is a smoothing parameter that affects the structure of the min-cut surface. In our experiments we found $\eta = 0.01$ to work well in practice. The capacity of source and sink edges are given as $c(s,v) = c(u,t) = \infty$ since they are assumed to allow maximum flow.

*D. From the maximum flow to best match surface*

Recalling that the problem is defined as to find the maximum quantity of flow $F$ from source vertex $s$ to sink vertex $t$, which depends mainly on the capacities of the edges $E$. Edges with lower capacities are more likely to be saturated by the maximum flow. Inversely, edges with higher capacities are unlikely to be saturated. In other words, it is the problem of finding the min-cut of the graph $G$, or the set of saturated edges that have minimal capacities.

Considering that the flow $F$ is a function of the capacities of edges $E$, we can write

$$F(c(E)) = F(c(u,v)), \quad (6)$$

then, the maximum flow is defined as

$$\hat{F} = \max F(c(u,v)), \quad (7)$$

which is solved by finding the set of saturated edges $(u', v')$ using the Goldberg algorithm. Reader is referred to [40], [41] for more details on the implementation of the Goldberg algorithm. It is worth noting that the capacity derived in Eq.(4), which is the average of the matching cost of the two nodes, is reported in the literature to be a formula that works well in solving matching problems [42].

Once a maximum flow is found through the graph $G = (V, E)$, it results in saturation of edges with lower capacities. These edges are represented by the min-cut $C$, which is formed in such a way so as to minimize the sum of these edge capacities. This cut therefore presents the optimal way of separating the source and the sink for the particular cost function. It gives, in turn, the surface representing best matches, see Fig. 4(a), from each of the reference sequences for a given frame in the test sequence.

Moreover, it can be proved that for every $(i, j)$ couple in the graph $G = (V, E)$, there exists at least one $k$ such that the edge $(i, j, k) - (i, j, k+1)$ is a part of cut $C$ [42]. In case we also get edges $(i, j, k) - (i, j+1, k)$ or $(i, j, k) - (i+1, j, k)$ to be a part of cut $C$, we take the vertex $(i, j, k)$ with the lowest value of $cost(u)$ that satisfies the above condition, to be a best match for the $j$th frame in the test sequence from the $i$th reference sequence. Similarly, we get a best match from each of the reference sequences, thus creating a best match surface. This gives us a total of $N$ frames representing best local matches from different reference sequences, for each frame in the test sequence. These frames generally correspond to the same physical place in the world, and thus have more or less similar values of $cost(u)$. But still, to get the best global match out of these $N$ frames and reject any possible outliers, the frame having minimum value of $cost(u)$ is chosen.

There are many algorithms which can be used to solve the maximum-flow problem [43]. For the work presented in this paper, we used the Goldberg algorithm, which uses the generic method known as $preflow-push$ [40], [41]. It has a time complexity of $O(\mathcal{D}^2\sqrt{E})$ where $\mathcal{D}$ and $E$ are the number of nodes and edges respectively.

## V. Experimental results

The experimental evaluation aims at supporting the proposed method for connecting visual experiences. The conducted experiments: (a) analyze the effect of the displacement parameter $k_{max}$, (b) show how the matching performance is improved when matching is done with respect to multiple image sequences, and finally (c) show the a comparison of our algorithm to the state-of-the-art image sequence matching methods, *i.e.* SeqSLAM [8] and ABLE-M [27]. Several datasets are used in our evaluation, three

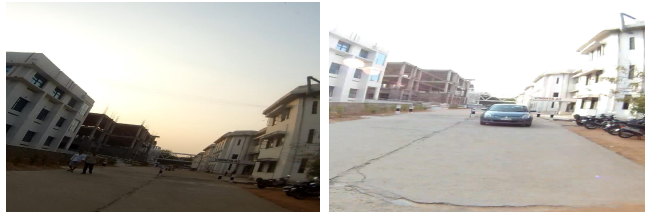

Fig. 5. An example of a difficult situation for place recognition in the IIIT-H dataset due to sunlight condition.

publicly available data sets and one of them is recorded in IIIT-H Campus.

### A. Data sets, ground truth, and evaluation

The datasets chosen allow us to test our proposal on the image sequences of places which are visibly different at different times. St. Lucia [44] and IIIT-H dataset [45] show changes along the day, Garden Point Walking dataset[1] exposes changes along the day and night with pose change, right and left angle of view, and the Nordland dataset [26] shows changes along the seasons. Each dataset has its own set of challenges and has been used to present the various features of the algorithm.

While the IIIT-H dataset has its various sequences varying in sync a lot, the Nordland dataset has completely synced sequences. This factor plays an important role in choosing a suitable value of displacement parameter $k_{max}$, as will be seen in Section V-B. These two datasets have been used to show the effect of the $k_{max}$ parameter. Garden Point Walking dataset, with its challenging conditions due to pose variance and extreme lighting conditions, has been used to compare the matching performance with SeqSLAM [8]. Finally, St. Lucia dataset, which contains video sequences recorded under varying illumination and shadow effects, has been used to further corroborate the advantage of using multiple sequences for life-long visual localization.

Our algorithm is mainly evaluated based on precision-recall curves obtained after getting a best match frame amongst all the training sequences for each test frame. We calculate precision as $\frac{TP}{TP+FP}$ and recall as $\frac{TP}{TP+FN}$. A threshold $t$ is used to separate positive matches from negative matches based on the $cost(u)$ value of the match $u$ between the test frame and the best frame found from the reference sequences.

$$\text{match} = \begin{cases} positive, & \text{if } cost(u) \leq t \\ negative, & \text{otherwise} \end{cases} \quad (8)$$

A positive match is considered as a $TP$ if the match found and the match provided by the ground truth differ up to a certain limit. This has been specified for different datasets. If an image pair is not within the specified range, it is considered as an $FP$. All undetected matches are considered $FN$.

---
[1]https://wiki.qut.edu.au/display/cyphy/
Day+and+Night+with+Lateral+Pose+Change+Datasets

TABLE I
COMPARISON OF OUR ALGORITHM FOR DIFFERENT VALUES
OF $k_{max}$ USING IIIT-H DATASET

| Threshold Distance (m) | Mean Error (m) | Percentage Error | Variance (m) | $k_{max}$ |
|---|---|---|---|---|
| 7  |        | 63.8 |         |     |
| 10 | 9.4858 | 39.7 | 23.1984 | 40  |
| 15 |        | 11.7 |         |     |
| 7  |        | 57.7 |         |     |
| 10 | 8.6064 | 34.1 | 17.2743 | 70  |
| 15 |        | 5.6  |         |     |
| 7  |        | 56.4 |         |     |
| 10 | 7.8319 | 26.8 | 13.2422 | 100 |
| 15 |        | 1.7  |         |     |

*B. Influence of the $k_{max}$ parameter*

One of the parameters that controls the behavior of our algorithm is $k_{max}$, which is presumably the most influential one. It controls the temporal length of the image sequences that are considered for matching. This parameter is highly dependent on the synchronization between different sequences of the dataset. For a dataset where the different runs vary a lot in sync, a higher value of $k_{max}$ will produce better matches than lower values, which could miss essential matches. To demonstrate this, we perform extensive experiments on the IIIT-H Campus dataset. This dataset contains several video sequences of the same route where illumination changes and shadow effect make it challenging for matching, as shown in Fig 5. The route has been traversed multiple times during the day. For the experiments, we down-sampled the dataset to 10fps and took the first 1000 frames for testing.

The performance of the algorithm for different values of $k_{max}$ can be seen in Table I. The first three runs were used for training and the fourth run was used for testing. We can see how increasing value of $k_{max}$ parameter improves the results in terms of mean GPS error and variance. This happens because the frames of different runs from this dataset are not in sync and increasing the value of $k_{max}$ allows for a better match to be found. This however, happens at the cost of efficiency since increasing the value of $k_{max}$ increases the number of nodes in the flow network. Mean error and percentage error are computed as in [10]. The mean error is calculated with respect to noisy GPS readings and is not related to the $TP$ threshold. Threshold distance for $TP$ affects only the percentage error, which denotes the percentage of test sequence frames having GPS error greater than the threshold distance.

On the other hand, for a frame-accurate dataset, a low value of $k_{max}$ would be preferred, as higher values would tend towards finding a global match. This can be further corroborated by experiments run on the Nordland dataset. The Nordland dataset is ∼3000 km in length and is a very challenging dataset in terms of life-long visual localization evaluation. It contains four sequences of a train ride that vary a lot mainly due to seasonal changes of appearance, besides other problematic aspects, as depicted in Fig 6(a). For the experiments, the dataset was down-sampled to 1fps and the first 1200 frames were used for testing. A match was considered to be a $TP$ if the positive match and the ground truth differed by up to one image within the sequence, as mentioned in [26].

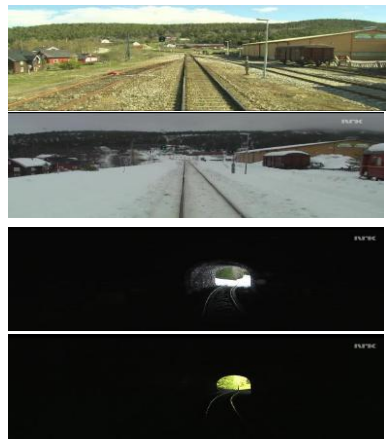

(a)

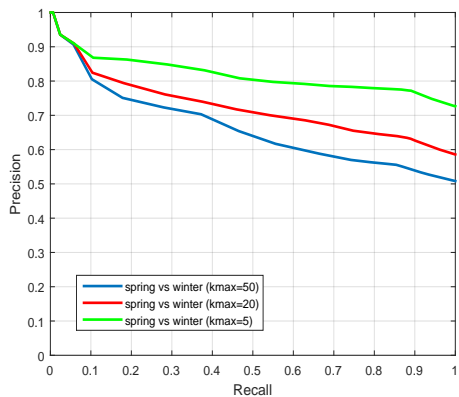

(b)

Fig. 6. Two examples from the Nordland dataset representing difficult situation for place recognition due to weather condition and lack of appearance shown in (a). In (b), performance comparison of our algorithm using precision recall curves between the Nordland sequences corresponding to two seasons, for different values of $k_{max}$.

For the IIIT-H Campus dataset, we showed the positive effect of increasing $k_{max}$ by showing comparative results for different values of it. We now present results on the Nordland dataset, whose frames are in complete sync. As depicted in Fig. 6(b), it can be clearly seen that in this case, decreasing the value of $k_{max}$ actually increases the performance of the algorithm. The increase in precision at 100% recall is approximately 10% and 22% for the first and second decrement respectively.

*C. Effect of multiple representations*

It is often challenging to detect matches from a single reference sequence, especially when considering the case of life-long visual localization where environment conditions are drastically changing and the image sequences aren't always viewpoint invariant. The lack of sync between the test and the reference sequence, which can be caused by stops made or speed of robot or vehicle, also produces difficulties in matching, This subsection shows how our algorithm efficiently exploits multiple image sequences for finding a best match, and how it is better than considering only a single reference sequence. The famous St. Lucia dataset is used for this purpose. It contains ten video sequences recorded in the year 2009 over a route where shadowing effects and varied illumination changes take place when traversed at different times of the day, as shown in Fig 7(a). We use six of those ten sequences for our experiments and label them A, B, C, D, E, and F. They were recorded at different dates and times [44]. For the experiments, first 7200 frames holding every 8th frame were used for testing. A match was considered to be a $TP$ if the distance between GPS coordinates of a positive match and the ground truth differed by up to 15 metres. Value of $k_{max} = 50$ was used.

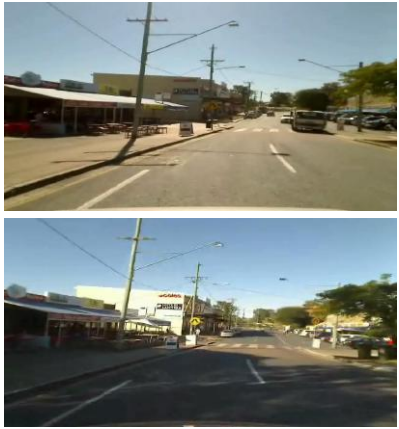

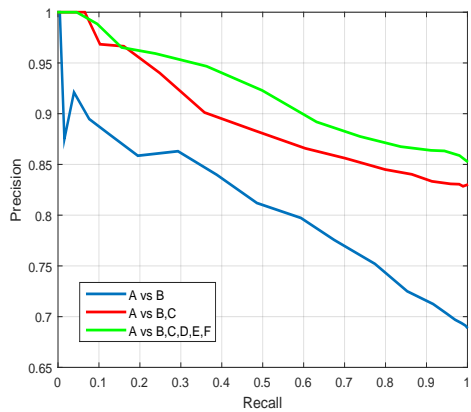

(a)

(b)

Fig. 7. An example from the St. Lucia dataset representing difficult situation for place recognition due to shadow and illumination effects at different times of the day is shown in (a). In (b), performance comparison of our algorithm using precision recall curves between the sequences of St. Lucia dataset for different number of training sequences. Threshold value of $TP$ is 15 metres.

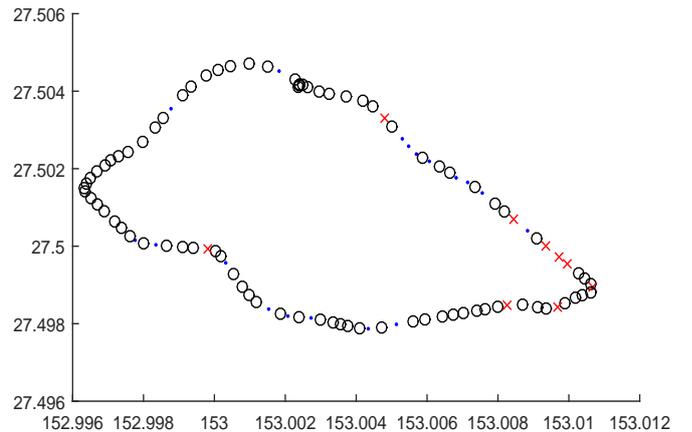

Fig. 8. Ground truth plot for the St. Lucia dataset using our algorithm. The black circles and red cross represent true (TP) and false positives (FP) respectively. Blue dots represent false negatives (FN).

We show the advantage of using multiple reference sequences for matching through a performance comparison using P-R curves in Fig 7(b). Here sequence A is used for testing against different combinations of reference sequences. When you compare the blue and the red curves, which represent matching results against one and two reference sequences respectively, the improvement in precision is approximately 15% at 100% recall. This is because due to several challenges such as appearance, vehicle speed and sync, not every frame in A is able to find a suitable match in B. But when C is added as a reference to the flow network, it serves as a source of "extra information" and provides A with an alternative for finding a suitable match in areas where B becomes challenging. When we further add D,E,F as references, we get further improvement in precision, but not as much as in the previous case. This also makes sense practically, since this "extra information" will get saturated eventually.

To further corroborate the above claim, we show an example of the result produced by our algorithm in Fig. 9 for the case where sequence A was used for testing and the rest were used as references. This example shows that even if some of the reference sequences are challenging in appearance, sync, and vehicle speed, which result in bad matches from sequences like B and D, a good match can be found from other references, like C, E and F, without increasing the complexity of the algorithm. This shows how multiple reference sequences help in overcoming each others' challenges for matching. The performance of the algorithm is also visualized in Fig. 8 using ground truth plot for St. Lucia dataset at 70% recall for the case where all reference sequences were used for matching.

### D. Comparison with SeqSLAM and ABLE-M

We now present results comparing the performance of our algorithm with SeqSLAM [8] and ABLE-M [27] on the Garden Point Walking dataset. This dataset contains three sequences of a single route through the Gardens Point Campus, Queenlands University of Technology. It is a challenging dataset given the day and night nature of the sequences, along with pose changes, as depicted in

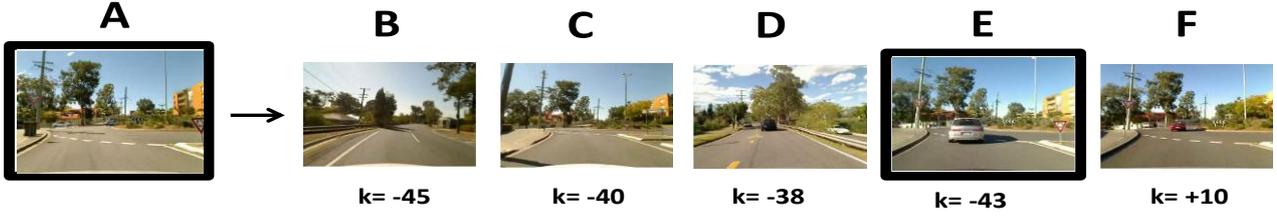

Fig. 9. An example of multiple sequences matching across six sequences of the St. Lucia dataset. Best local match form each reference sequence has been shown, along with value of the corresponding displacement parameter $k$. Match from reference sequence E represents global best match.

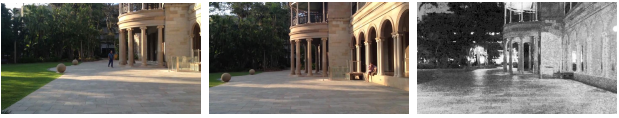

Fig. 10. Examples from the Garden Point Walking dataset representing difficult situations for place recognition. The images represent pose variance as well as day and night difference at the same place

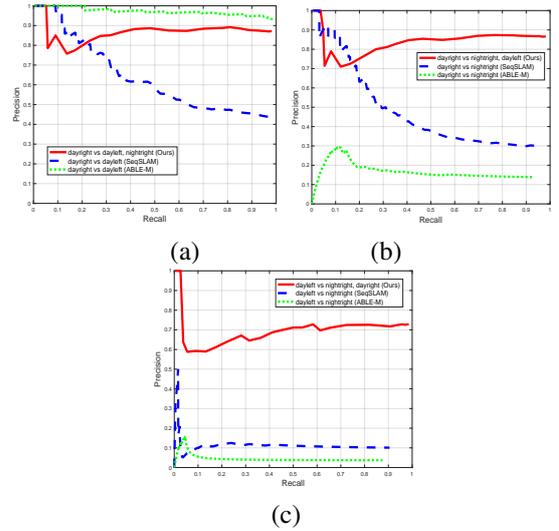

Fig. 11. Precision-recall curves comparing the performance of our algorithm with SeqSLAM and ABLE-M using different combination of training and test sequences of Garden Point Walking dataset.

Fig 10. For comparison purpose, we have used the source codes provided by OpenSeqSLAM [26] for the evaluation of SeqSLAM, and OpenABLE [14] for the evaluation of ABLE-M. We use their standard configurations, with sequence length parameter as 10 and 20 for both SeqSLAM and ABLE-M respectively, along with illumination invariance for ABLE-M. This subsection presents results corresponding to different scenarios, that are a combination of varied test and training sequences, details of which can be found in Fig. 11. A match was considered to be a $TP$ if the positive match and the ground truth differed by up to three images within the sequence. Value of $k_{max} = 5$ is used.

As can be gleaned from the curves in Figs. 11(a, b, c), our algorithm outperforms SeqSLAM in all scenarios with recall value greater than 20%. We also note here that our method is not able to perform better than ABLE-M in Fig. 11(a). This is primarily due to the kind of features used for matching, and we later show in Section V-E how using better features helps us perform favorably against ABLE-M for this setting. That being said, our method significantly outperforms ABLE-M in Figs. 11(b,c) in which lighting condition change from day to night. Since we consider multiple reference sequences for finding a best match, we get multiple match cases for consideration. So, even if a suitable match is not found from one of the sequences due to high pose variance and/or lighting condition change, there is always the possibility of finding a match from another sequence which might not have such issues. SeqSLAM and ABLE-M consider only one reference sequence at a time and so cannot take advantage of this fact. Their poor performance is especially evident in the case of Day-left vs Night-right in Fig. 11(c), where both viewpoint as well as lighting conditions are different. Our algorithm in this particular case performs far better because it is able to exploit information and pick the best out of both the reference sequences.

We also show an example of the results produced by our algorithm in Fig 12, where Day-left sequence was used for testing and rest for reference purposes, along with the output from SeqSLAM and ABLE-M. Due to pose variance and lighting change, a good match could not be found in the sequence Night-right. But since we also considered sequence Day-right for matching, which does not have such problem, a more suitable match was found for the same frame in the test sequence. SeqSLAM and ABLE-M on the other hand did not have such an option, and had to settle for a poor match as shown.

### E. Discussion

We discuss in this section some aspects of the proposed algorithm and the assumptions of the experimental evaluation. The major assumption in our algorithm is that the search for a match with the $j$th test frame is limited in the range $[-k_{max}, +k_{max}]$. In other words, all reference sequences are assumed to be recorded using similar speed profiles. This assumption however, is not a prerequisite for the working of our algorithm, since highly unsynchronised data can be dealt with by sampling and increasing the value of $k_{max}$ as shown in Section V-B. In case there is no priori information about the value of $k_{max}$, it can be selected equal to the half the length of the shortest reference sequence. For example, the assumption that experiences are recorded sequentially like in [4], [18] simplifies the problem to sequence to sequence matching. This assumption is natural since a new visual experience is recorded for the same place every time the vehicle visits the route again. The newly recorded experience is applied as input to the alignment algorithm to connect it with the previously available experiences. When only one previous experience is available, the matching is done as one sequence to one sequence. The value of $k_{max}$ is selected equal to half the length of the shorter experience. Then, a rough estimate of the value of $k_{max}$ can be obtained and used when more experiences are available. However, choosing an appropriate value of $k_{max}$ is beyond the scope of the current work, and we plan to explore it in future work.

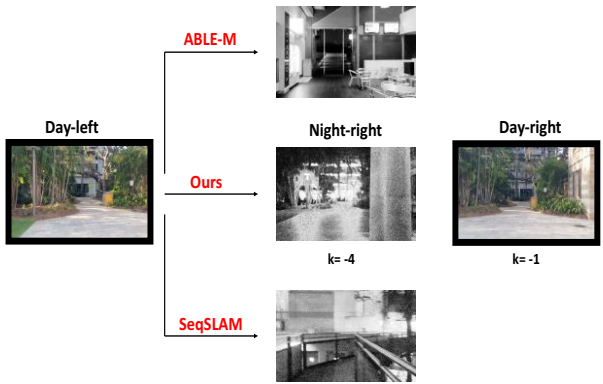

Fig. 12. An example of multiple sequences matching using our algorithm across the three sequences of the Garden Point Walking dataset, along with matching output from SeqSLAM and ABLE-M. Best local match from each reference sequence has been shown along with the value of the corresponding displacement parameter $k$. Match from reference sequence Day-right represents best global match.

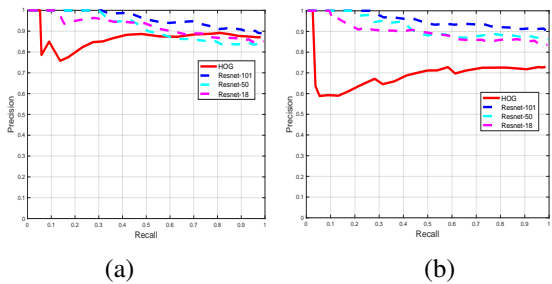

Fig. 13. Precision-recall curves comparing the performance of our algorithm using different features on the Garden Point Walking dataset. In (a), Day-right sequence is used for testing vs Day-left and Night-right as references. Day-left sequence is used for testing vs Night-right and Day-right as references in (b).

TABLE II
COMPARISON OF PROCESSING TIMES FOR THE IMAGE MATCHING STAGES OF SEQSLAM, ABLE-M AND OUR METHOD.

| No. of reference sequences | No. of images to match | SeqSLAM | ABLE-M | Ours |
|---|---|---|---|---|
| 1 | 400 | .09s | .01s | .37s |
| 2 | 600 | N/A | N/A | .64s |

Although is this paper we make us of HOG features for image matching, the proposed framework can also be used with other discriminating features. In Fig 13, we use precision-recall curves to show a comparison between HOG features and features extracted from the last layer of different deep residual networks [38] which are pretrained on the ImageNet dataset [46]. Although the features are different, we use the same similarity measure for computing matching costs, as described in Section III-A. We show this for two settings of the Garden Point Walking dataset, which are present in Figs 11(a,c) as well. As can be gleaned from the figures, the deep features significantly outperform HOG. It is interesting to note here that our method, with Resnet-101 features in Fig 13(a), performs equally well at higher recall rates and slightly better at lower recall rates than ABLE-M under the same setting in Fig 11(a). This clearly shows that our method provides a generic framework for matching across multiple sequences, with the flexibility of task specific features and similarity metric holding the potential for further improvement in results. However, exploring such features and metric is not the focus of this work, and we plan to explore it in future work.

The tolerance in speed is directly reflected in the value of $k_{max}$. Even though the algorithm does not explicitly consider the sequential nature of the route, it is globally considered since the best match of the frame $x_j$ is always assumed to be in the range $[y_i - k_{max}, y_i + k_{max}]$. We have provided detailed analysis on how the value of $k_{max}$ affects the performance of our algorithm for difference in speed or sync. Besides accuracy in performance, value of $k_{max}$ also affects the efficiency of our algorithm. This is because in deciding the number of frames to be considered for matching, $k_{max}$ provides the number of nodes in the 3D flow network.

Even when considering multiple reference sequences for matching, our algorithm is reasonably fast. In Table II, we present the processing times for image matching for different methods corresponding to the results in Section V-D. For a fair comparison, we show processing time of our algorithm when both one and two reference sequences are used. We note here that our algorithm has a higher processing time than both SeqSLAM and ABLE-M as shown in Table II. However, this is acceptable since the alignment is not assumed to be a real time process. Image matching in SeqSLAM and ABLE-M is equivalent to finding the min-cut surface in our case, and therefore, the processing time shown in Table II does not include the time required for maximum flow computation. However, even after including maximum flow computation part, our algorithm takes .67s and 1.38s for the case with one and two reference sequences respectively, which is reasonable. In tests, we use a standard 2.7 GHz Intel Core i7 PC.

Given the fact that ABLE-M outperforms our method in Fig 11(a) and is much faster as shown in Table II, one could argue that a better solution would be to run ABLE-M on all available sequences and choose the best results. But, we would like to point out here that we get equally good results by changing the features used for matching in Fig 13(a), and thus argue that it's rather a matter of choosing the right features and metric for matching. Besides, the focus of this paper is not to outperform state of the art methods for visual localization, but to provide a generic framework for multiple image sequence alignment.

## VI. CONCLUSIONS

An algorithm that connects visual experiences using multiple image sequence matching is presented in this paper. The matching process is represented as a 3D maximum network flow problem. Our algorithm also presents an innovative way of checking sequential nature of matches not only within the same reference sequence but also between different reference sequences, with the help of the 6-connected edge structure of the flow network. We demonstrate the effectiveness of the algorithm via application to visual localization. The performance of our algorithm along with analysis of its most important features has been shown by testing over some of the most challenging available datasets. Some of them show changes over the day or even the seasons, while others show changes along the day and night with pose change, i.e. right and left angle of view.

The performance is evaluated using the precision-recall curves. The results show a high precision over a wide range of recall values. Furthermore, we are able to show how our multiple image sequence matching algorithm performs favorably against SeqSLAM and ABLE-M.

## REFERENCES

[1] F. Bonin-Font, A. Ortiz, and G. Oliver, "Visual navigation for mobile robots: A survey," *Journal of intelligent and robotic systems*, vol. 53, no. 3, p. 263, 2008.

[2] C. Cadena, L. Carlone, H. Carrillo, Y. Latif, D. Scaramuzza, J. Neira, I. Reid, and J. J. Leonard, "Past, present, and future of simultaneous localization and mapping: Toward the robust-perception age," *IEEE Transactions on Robotics*, vol. 32, no. 6, pp. 1309–1332, 2016.

[3] S. Lowry, N. Sunderhauf, P. Newman, J. J. Leonard, D. Cox, P. Corke, and M. J. Milford, "Visual place recognition: A survey," *IEEE Trans. Robot.*, 2016.

[4] W. Churchill and P. Newman, "Experience-based navigation for long-term localisation," *IJRR*, 2013.

[5] M. Dymczyk, S. Lynen, T. Cieslewski, M. Bosse, R. Siegwart, and P. Furgale, "The gist of maps-summarizing experience for lifelong localization," in *ICRA*, 2015.

[6] R. Arroyo, P. F. Alcantarilla, L. M. Bergasa, and E. Romera, "Are you able to perform a life-long visual topological localization?" *Autonomous Robots*, pp. 1–21, 2017.

[7] F. Bellavia, M. Fanfani, and C. Colombo, "Selective visual odometry for accurate auv localization," *Autonomous Robots*, vol. 41, no. 1, pp. 133–143, 2017.

[8] M. Milford and G. F. Wyeth, "Seqslam: Visual route-based navigation for sunny summer days and stormy winter nights," in *ICRA*, 2012.

[9] C. Linegar, W. Churchill, and P. Newman, "Work smart, not hard: Recalling relevant experiences for vast-scale but time-constrained localisation," in *ICRA*. IEEE, 2015, pp. 90–97.

[10] A. H. A. Hafez, M. Arora, K. M. Krishna, and C. V. Jawahar, "Learning multiple experiences useful visual features for active maps localization in crowded environments," *Advanced Robotics*, 2016.

[11] T. Naseer, B. Suger, M. Ruhnke, and W. Burgard, "Vision-based markov localization for long-term autonomy," *Robotics and Autonomous Systems*, vol. 89, pp. 147–157, 2017.

[12] R. Arroyo, P. F. Alcantarilla, L. M. Bergasa, and E. Romera, "Fusion and binarization of cnn features for robust topological localization across seasons," in *Intelligent Robots and Systems (IROS), 2016 IEEE/RSJ International Conference on*. IEEE, 2016, pp. 4656–4663.

[13] N. Sünderhauf, S. Shirazi, F. Dayoub, B. Upcroft, and M. Milford, "On the performance of convnet features for place recognition," in *Intelligent Robots and Systems (IROS), 2015 IEEE/RSJ International Conference on*. IEEE, 2015, pp. 4297–4304.

[14] R. Arroyo, P. F. Alcantarilla, L. M. Bergasa, and E. Romera, "Openable: An open-source toolbox for application in life-long visual localization of autonomous vehicles," in *Intelligent Transportation Systems (ITSC), 2016 IEEE 19th International Conference on*. IEEE, 2016, pp. 965–970.

[15] C. Stachniss and W. Burgard, "Mobile robot mapping and localization in non-static environments," in *aaai*, 2005, pp. 1324–1329.

[16] A. Ranganathan, S. Matsumoto, and D. Ilstrup, "Towards illumination invariance for visual localization," in *Robotics and Automation (ICRA), 2013 IEEE International Conference on*. IEEE, 2013, pp. 3791–3798.

[17] P. Hansen and B. Browning, "Visual place recognition using hmm sequence matching," in *IROS*. IEEE, 2014, pp. 4549–4555.

[18] T. Naseer, M. Ruhnke, C. Stachniss, L. Spinello, and W. Burgard, "Robust visual slam across seasons," in *IROS*, 2015.

[19] O. Vysotska, T. Naseer, L. Spinello, W. Burgard, and C. Stachniss, "Efficient and effective matching of image sequences under substantial appearance changes exploiting gps priors," in *ICRA*, 2015.

[20] M. Cummins and P. Newman, "FAB-MAP: Probabilistic Localization and Mapping in the Space of Appearance," *IJRR*, 2008.

[21] A. Glover, W. Maddern, M. Warren, S. Reid, M. Milford, and G. Wyeth, "Openfabmap: An open source toolbox for appearance-based loop closure detection," in *ICRA*, 2012.

[22] N. Carlevaris-Bianco and R. M. Eustice, "Learning temporal co-observability relationships for lifelong robotic mapping," in *IROS Workshop on Lifelong Learning for Mobile Robotics Applications*, 2012, p. 15.

[23] L. Murphy, S. Martin, and P. Corke, "Creating and using probabilistic costmaps from vehicle experience," in *IROS*, 2012.

[24] E. Johns and G.-Z. Yang, "Generative methods for long-term place recognition in dynamic scenes," *IJCV*, vol. 106, no. 3, pp. 297–314, 2014.

[25] A. Hafez, M. Singh, K. M. Krishna, and C. Jawahar, "Visual localization in highly crowded urban environments," in *IROS*, 2013.

[26] N. Sünderhauf, P. Neubert, and P. Protzel, "Are we there yet? challenging seqslam on a 3000 km journey across all four seasons," in *ICRA*, 2013.

[27] R. Arroyo, P. F. Alcantarilla, L. M. Bergasa, and E. Romera, "Towards life-long visual localization using an efficient matching of binary sequences from images," in *Robotics and Automation (ICRA), 2015 IEEE International Conference on*. IEEE, 2015, pp. 6328–6335.

[28] T. Naseer, L. Spinello, W. Burgard, and C. Stachniss, "Robust visual robot localization across seasons using network flows," in *AAAI*, 2014.

[29] K. Mikolajczyk and C. Schmid, "A performance evaluation of local descriptors," *PAMI*, 2005.

[30] D. G. Lowe, "Distinctive image features from scale-invariant keypoints," *International journal of computer vision*, vol. 60, no. 2, pp. 91–110, 2004.

[31] W. Burger and M. J. Burge, "Scale-invariant feature transform (sift)," in *Digital Image Processing*. Springer, 2016, pp. 609–664.

[32] M. Calonder, V. Lepetit, M. Ozuysal, T. Trzcinski, C. Strecha, and P. Fua, "Brief: Computing a local binary descriptor very fast," *IEEE Transactions on Pattern Analysis and Machine Intelligence*, vol. 34, no. 7, pp. 1281–1298, 2012.

[33] N. Dalal and B. Triggs, "Histograms of oriented gradients for human detection," in *CVPR*, 2005.

[34] D. Geronimo, A. M. Lopez, A. D. Sappa, and T. Graf, "Survey of pedestrian detection for advanced driver assistance systems," *IEEE transactions on pattern analysis and machine intelligence*, vol. 32, no. 7, pp. 1239–1258, 2010.

[35] V. Ramakrishnan, A. K. Prabhavathy, and J. Devishree, "A survey on vehicle detection techniques in aerial surveillance," *International Journal of Computer Applications*, vol. 55, no. 18, 2012.

[36] L. Chen and N. Hassanpour, "Survey: How good are the current advances in image set based face identification?–experiments on three popular benchmarks with a naïve approach," *Computer Vision and Image Understanding*, vol. 160, pp. 1–23, 2017.

[37] C. Bila, F. Sivrikaya, M. A. Khan, and S. Albayrak, "Vehicles of the future: A survey of research on safety issues," *IEEE Transactions on Intelligent Transportation Systems*, vol. 18, no. 5, pp. 1046–1065, 2017.

[38] K. He, X. Zhang, S. Ren, and J. Sun, "Deep residual learning for image recognition," in *Proceedings of the IEEE conference on computer vision and pattern recognition*, 2016, pp. 770–778.
[39] M. Milford, "Visual route recognition with a handful of bits," *Proc. 2012 Robotics: Science and Systems VIII*, pp. 297–304, 2012.
[40] R. K. Ahuja, M. Kodialam, A. K. Mishra, and J. B. Orlin, "Computational investigations of maximum flow algorithms," *Euro. Jour. of Operational Research*, 1997.
[41] A. V. Goldberg, "The partial augment–relabel algorithm for the maximum flow problem," in *Algorithms-ESA*, 2008.
[42] S. Roy and I. J. Cox, "A maximum-flow formulation of the n-camera stereo correspondence problem," in *CVPR*, 1998.
[43] T. H. Cormen, *Introduction to algorithms*. MIT press, 2009.
[44] A. J. Glover, W. P. Maddern, M. J. Milford, and G. F. Wyeth, "Fabmap+ ratslam: appearance-based slam for multiple times of day," in *ICRA*, 2010.
[45] K. Vishal, C. V. Jawahar, and V. Chari, "Accurate localization by fusing images and GPS signals," in *CVPR Workshops*, 2015.
[46] J. Deng, W. Dong, R. Socher, L.-J. Li, K. Li, and L. Fei-Fei, "Imagenet: A large-scale hierarchical image database," in *Computer Vision and Pattern Recognition, 2009. CVPR 2009. IEEE Conference on*. IEEE, 2009, pp. 248–255.